%% file: main.tex
\documentclass{applemlr}
\input{math_commands.tex}

\usepackage{url}
\usepackage{graphicx}
\usepackage{booktabs}
\usepackage[table]{xcolor}
\usepackage{natbib}
\usepackage{pifont}
\usepackage{multirow}
\usepackage{wrapfig}
\usepackage{listings}
\usepackage{mdframed}
\usepackage{enumitem}
\usepackage{amssymb} % For symbols like \downarrow
\definecolor{darkgreen}{HTML}{20662A} % A nice dark green color
\usepackage{graphicx}
\usepackage{subcaption}
\usepackage[textwidth=18mm]{todonotes}
\usepackage{ulem}  
\usepackage{xcolor} 
\usepackage{siunitx}
\usepackage{xspace}
\usepackage{longtable}
\usepackage{color}
\usepackage[dvipsnames]{xcolor}
\definecolor{JalapenoRed}{RGB}{183,21,64}
\definecolor{Belize}{RGB}{41,128,185}
\definecolor{Amour}{RGB}{238,82,83}
\definecolor{textgray}{HTML}{6E6E73}
\usepackage{colortbl}
\usepackage{makecell}

\makeatletter
\newcommand\applefootnote[1]{%
  \begingroup
  \renewcommand\thefootnote{}%
  \renewcommand\@makefntext[1]{\noindent##1}%
  \footnote{#1}%
  \addtocounter{footnote}{-1}%
  \endgroup
}
\makeatother

\definecolor{tableheader}{RGB}{70,130,180}
\definecolor{oddrow}{RGB}{245,245,245}
\definecolor{evenrow}{RGB}{255,255,255}

\newcommand{\code}[1]{\texttt{\footnotesize\color{black!80}#1}}

\newcommand{\affh}{\textsuperscript{\textcolor{purple}{\textbf{h}}}}
\newcommand{\affa}{\textsuperscript{\textcolor{purple}{\textbf{a}}}}

\title{UltraCUA: A Foundation Model for Computer Use Agents with Hybrid Action}

\author{
\parbox{\textwidth}{
Yuhao Yang\affh, Zhen Yang\affa, Zi-Yi Dou\affa, Anh Nguyen\affa, Keen You\affa, Omar Attia\affa,
Andrew Szot\affa, Michael Feng\affa, \\ Ram Ramrakhya\affa, Alexander Toshev\affa$^\dagger$, Chao Huang\affh$^\dagger$,
Yinfei Yang\affa$^\dagger$, Zhe Gan\affa$^\dagger$
}}
\affiliation{\affa Apple}
\affiliation{\affh The University of Hong Kong}
\contribution{$^\dagger$Senior Authors}

\newcommand{\paratitle}[1]{\noindent\textbf{#1}}
\newcommand{\model}{UltraCUA\xspace}

\lstdefinestyle{trajectorystyle}{
    backgroundcolor=\color{backcolour},
    basicstyle=\ttfamily\footnotesize,
    breaklines=true, % This automatically wraps long lines
    frame=single,
    frameround=tttt,
    rulecolor=\color{black!30}
}

\definecolor{codegreen}{rgb}{0,0.6,0}
\definecolor{codegray}{rgb}{0.5,0.5,0.5}
\definecolor{codepurple}{rgb}{0.58,0,0.82}
\definecolor{backcolour}{rgb}{0.98,0.98,0.98}

\lstdefinestyle{mystyle}{
    backgroundcolor=\color{backcolour},   
    commentstyle=\color{codegreen},
    keywordstyle=\color{magenta},
    stringstyle=\color{codepurple},
    basicstyle=\ttfamily\footnotesize,
    breakatwhitespace=false,         
    breaklines=true,                 
    captionpos=b,                    
    keepspaces=true,                 
    showspaces=false,                
    showstringspaces=false,
    showtabs=false,                  
    tabsize=2,
    frame=single, % Adds a simple, single-line frame
    frameround=tttt, % Slightly rounded corners
    rulecolor=\color{black!30} % A light gray for the frame
}

\abstract{
Computer-use agents face a fundamental limitation. They rely exclusively on primitive GUI actions (click, type, scroll), creating brittle execution chains prone to cascading failures. While API-driven agents harness rich capabilities through structured interfaces and tools, computer-use agents remain constrained to low-level visual interactions. We present UltraCUA, a foundation model that transcends this limitation through hybrid action—seamlessly unifying primitive GUI operations with high-level tool execution. Our innovation rests on four critical advances. First, an automated pipeline extracts and scales tool capabilities from software documentation and code repositories. Second, a synthetic data engine produces 17,000+ verifiable tasks capturing real-world computer-use complexity. Third, comprehensive hybrid action trajectory collection incorporates both GUI primitives and strategic tool calls. Fourth, a two-stage training methodology combines supervised fine-tuning with online reinforcement learning, enabling intelligent action selection between GUI and API. Evaluation with our 7B and 32B UltraCUA models reveals transformative performance gains. On OSWorld, UltraCUA achieves 22\% relative improvement while executing 11\% faster than existing approaches on average. Cross-domain validation on WindowsAgentArena demonstrates robust generalization with 21.7\% success rate, surpassing Windows-trained baselines. The hybrid action paradigm proves essential, reducing error propagation while improving execution efficiency. This work establishes a scalable paradigm bridging primitive GUI interactions and high-level tool intelligence, enabling more resilient and adaptable computer use agents for diverse environments and complex real-world tasks.
}
\date{\sffamily October 21, 2025}

\begin{document}
\maketitle

\vspace{-0.1in}
\begin{figure*}[h!]
    \centering
    \begin{subfigure}[b]{0.64\textwidth}
        \centering
        \includegraphics[height=4cm]{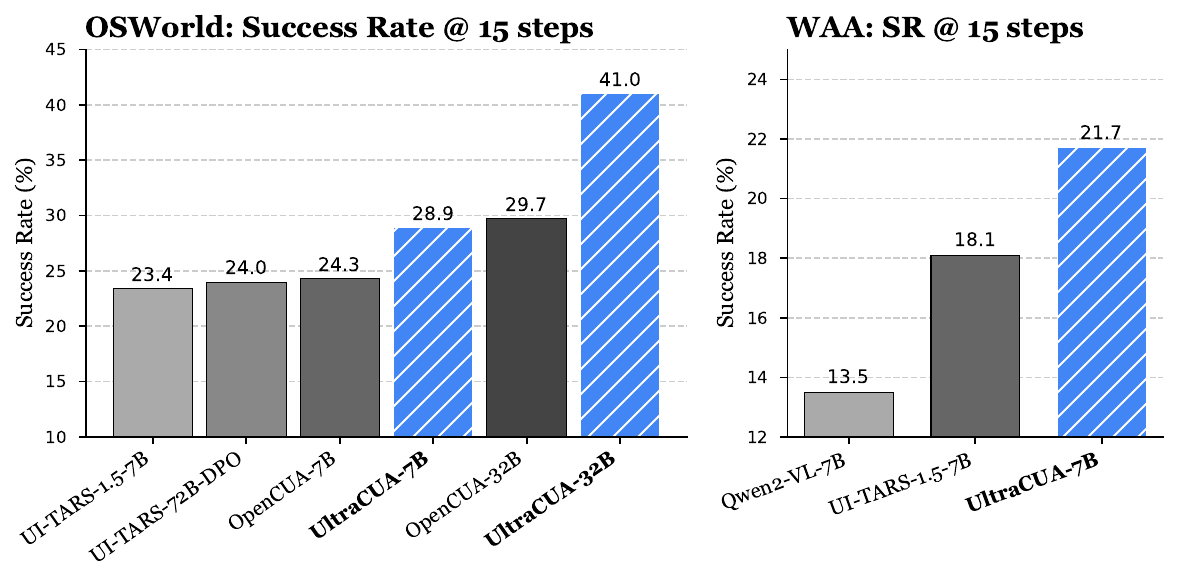}
        \caption{OSWorld and WindowsAgentArena}
        \label{fig:osworld_results}
    \end{subfigure}
    % \begin{subfigure}[b]{0.32\textwidth}
    %     \centering
    %     \includegraphics[height=3.5cm]{resources/teaser_waa.pdf}
    %     \caption{WindowsAgentArena}
    %     \label{fig:windows_results}
    % \end{subfigure}
    \hspace{0.02in}
    \begin{subfigure}[b]{0.33\textwidth}
        \centering
        \includegraphics[height=4cm]{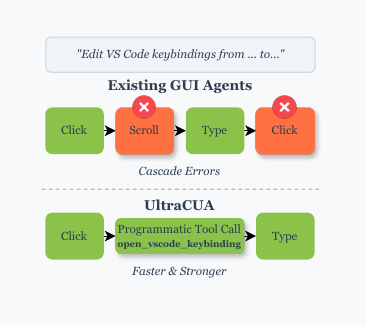}
        \caption{Existing Agents v.s. \model}
        \label{fig:comparison_gui_ultracua}
    \end{subfigure}
    \vspace{-0.05in}
    % \caption{(a) \model's performance on OSWorld and WindowsAgentArena; (b) Comparison between existing GUI Agents and \model. Pure low-level actions lead to cascade errors, while \model is faster and stronger.}
    \caption{(a) UltraCUA demonstrates superior performance on OSWorld and WindowsAgentArena benchmarks; (b) Comparison of execution paradigms between existing GUI agents and UltraCUA. Traditional agents using pure low-level actions suffer from cascading errors, while UltraCUA's hybrid approach achieves faster and more robust execution.}    \label{fig:teaser}
    \vspace{-0.15in}
\end{figure*}

\input{intro}
\input{method}
\input{exp}
\input{conclusion}

\section*{Acknowledgment}
The authors would like to thank Harsh Agrawal, Dongxu Li, and Yutong Dai for valuable suggestions and feedback.

\applefootnote{ \textcolor{textgray}{\sffamily Apple and the Apple logo are trademarks of Apple Inc., registered in the U.S. and other countries and regions.}}

\bibliography{ref}
\bibliographystyle{main}

\include{appendix}

\end{document}

%% file: math_commands.tex
\usepackage{amsmath,amsfonts,bm}

\def\eqref#1{equation~\ref{#1}}
\def\1{\bm{1}}

\DeclareMathAlphabet{\mathsfit}{\encodingdefault}{\sfdefault}{m}{sl}
\SetMathAlphabet{\mathsfit}{bold}{\encodingdefault}{\sfdefault}{bx}{n}

%% file: intro.tex
\section{Introduction}
Computer-use automation has emerged as a transformative capability for autonomous agents. It enables interaction with the vast ecosystem of desktop and web applications that define modern digital workflows \citep{hong2024cogagent, shaw2024pixels, zhang2023appagent}. However, current computer-use agents (CUAs) face a critical architectural limitation. They operate exclusively through primitive visual actions such as clicking, typing, and scrolling \citep{rawles2024androidworld, koh2024visualwebarena}. This constraint creates a profound performance chasm compared to API-driven agents that harness rich tool-based interfaces---MCP servers and specialized tools to accomplish complex tasks with efficiency \citep{qin2023toolllm, schick2023toolformer}.

The exclusive reliance on primitive actions introduces fundamental challenges that severely limit agent effectiveness. First, extended execution chains accumulate errors that propagate into catastrophic failures. A single misplaced click can completely derail complex workflows \citep{zheng2024gpt4vision, yan2023gpt4v}. Second, operations achievable through single tool calls require dozens of fragile GUI interactions. This creates severe performance bottlenecks. For instance, extracting data from multiple spreadsheets forces traditional CUAs to navigate complex menus, manually select individual cells, copy values, switch applications, and paste content. Each step represents a critical failure point. Conversely, an agent with native spreadsheet APIs could accomplish this task reliably with minimal operations. This efficiency disparity is striking. While agents leveraging tool-based interfaces consistently exceed 80\% success on challenging benchmarks like GAIA \citep{mialon2024gaia, zhang2025agentorchestra}, GUI-constrained computer-use agents remain fundamentally bottlenecked. This compels our unified approach that marries GUI universality with tool-based precision.

In this paper, we bridge this capability gap through \textbf{hybrid action}, a paradigm that recognizes the complementary strengths of both execution modalities. Our approach seamlessly integrates GUI primitives with high-level tool execution, unlocking the best of both worlds: the universal applicability of visual interactions and the efficiency of direct tool access. Rather than treating these paradigms as mutually exclusive alternatives, we enable intelligent orchestration that maximizes task success while minimizing execution overhead. Agents learn to strategically deploy tool calls when they offer clear efficiency advantages, while preserving GUI interactions for comprehensive coverage and granular control when tools are unavailable or insufficient.

To summarize, our technical contributions include:

\begin{itemize}[leftmargin=*]

\item \textbf{An automated pipeline for synthesizing tool capabilities} that transcends manual curation limitations \citep{qin2023tool, tang2023toolalpaca}. Unlike existing approaches that rely on fixed tool sets, our system dynamically extracts tools from software documentation, integrates open-sourced implementations, and employs coding agents to generate tools on demand. This enables scalable tool acquisition across diverse environments, producing hundreds of tools that form a critical foundation for hybrid action execution.

\item \textbf{A dual-pipeline synthetic data engine} that solves the fundamental challenge of verifiable task generation for computer-use agent training. Traditional approaches fail to reliably assess task completion in dynamic environments, creating unreliable training data. We develop two innovative pipelines that together produce 17,000+ tasks with guaranteed verification. The instruction-first pipeline leverages agent exploration to generate contextually relevant tasks, with dedicated evaluators ensuring execution validity. The evaluator-first pipeline reverses this process by collecting atomic verification functions from target environments, then reprogramming or composing them for complex assessment criteria. Tasks are generated to satisfy these pre-built evaluators, ensuring reliable completion detection. This approach addresses the verification bottleneck that has limited training computer-use agents at scale.

\item \textbf{A large-scale hybrid action trajectory collection} that bridges the critical gap between pure GUI datasets and tool-augmented execution. Existing computer-use datasets contain only primitive GUI action sequences. They lack demonstrations of strategic tool integration, severely limiting agent learning capabilities. We address this fundamental limitation by collecting 20,000+ successful trajectories using an multi-agent framework. Our approach combines a powerful planner (e.g., OpenAI o3) with a state-of-the-art grounding model (e.g., GTA1-7B~\citep{yang2025gta1}). The planner intelligently selects between tool calls and GUI actions based on efficiency considerations. The grounder ensures precise execution of chosen actions. This dataset uniquely captures the decision-making patterns required for seamless action mode transitions, enabling agents to learn when and how to leverage hybrid capabilities.

\item \textbf{An agentic foundation model with hybrid action intelligence} that transcends the limitations of GUI-only training. We develop models at two scales (7B and 32B) through our data and tool ecosystem. Our training employs a strategic two-stage methodology. First, supervised fine-tuning on high-quality hybrid trajectories teaches agents when to leverage tools versus GUI actions. Second, online reinforcement learning on verifiable synthetic tasks refines decision-making through on-policy feedback. This approach produces agents that strategically balance tool efficiency with GUI universality.
\end{itemize}

Experiments reveal the transformative impact of UltraCUA's hybrid action across diverse benchmarks and platforms. On OSWorld \citep{xie2024osworld}, our UltraCUA models deliver a compelling 22\% relative improvement over base models at 7B and 32B scales, averagely. In terms of cross-platform generalization, UltraCUA-7B achieves 21.7\% success on WindowsAgentArena~\citep{bonatti2024windows} without any Windows-specific training, surpassing models explicitly trained on Windows data. This demonstrates that UltraCUA's hybrid action capabilities fundamentally enhance agent adaptability rather than merely optimizing for specific environments. The consistent gains across model scales and operating systems establish UltraCUA as a paradigm for robust computer-use automation.
\vspace{-0.15in}

%% file: method.tex
\section{Methodology}

\begin{figure}[t!]
    \centering
    \includegraphics[width=\linewidth]{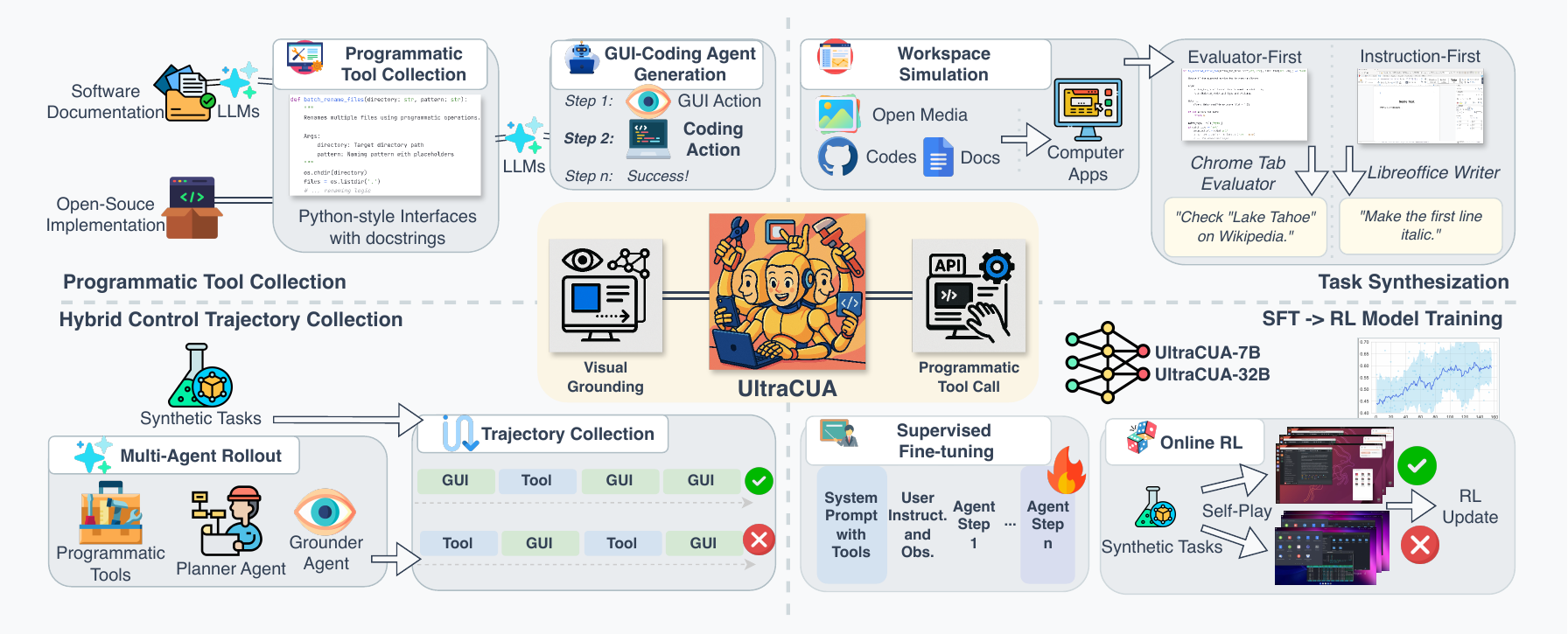}
    %\vspace{-0.3in}
    \caption{An overview of UltraCUA's design. UltraCUA adaptively switches between visual grounding and tool execution based on task requirements. This establishes the hybrid action mechanism that seamlessly integrates GUI primitives with high-level tool capabilities for enhanced computer-use automation across diverse environments.}
    \label{fig:framework}
    \vspace{-0.15in}
\end{figure}

Our methodology comprises three key components for developing a foundation CUA model with hybrid action. First, we build a comprehensive collection of programmatic tools through an automated extraction pipeline. Second, we design a dual-pipeline synthetic data engine that generates verifiable tasks for complex real-world computer use. Finally, we train our model via supervised fine-tuning on collected trajectories followed by online reinforcement learning on synthetic tasks.

\subsection{Automated Tool Collection for Hybrid Action}

The foundation of our approach is \textbf{hybrid action}—seamlessly integrating primitive GUI actions with high-level programmatic tools. We define a ``tool'' as a high-level interface encapsulating sequences of computer-use actions, typically implemented as Python functions, keyboard shortcuts, or combinations of primitive actions (e.g., type, key combinations)—but excluding actions requiring visual grounding like clicks. Each tool is exposed to the model through a Python function signature with descriptive docstrings specifying parameters and functionality.

While GUI-only agents suffer from cascading failures in lengthy action sequences, programmatic interfaces alone cannot handle all computer interactions. Our hybrid approach enables agents to leverage programmatic tools for efficiency when available, while retaining GUI actions to ensure generalization. To build a hybrid action space where tools cover diverse applications and usage scenarios, we developed an automated pipeline collecting hundreds of tools from the following three complementary sources. Tool details are also present in Appendix~\ref{app:tool}.

\textbf{Extraction from Software Documentation.} Application documentation contains expert knowledge, particularly keyboard shortcuts, that bypass tedious GUI sequences. For example, changing VS Code's color theme requires navigating \texttt{File → Preferences → Color Theme} with GUI actions. Our pipeline extracts the shortcut (\texttt{Ctrl+K, Ctrl+T}) from documentation and converts it into a programmatic tool: \texttt{vscode.set\_theme()}. This transforms fragile multi-step sequences into single, reliable operations.

\textbf{Integration of Open-Source Implementations.} We incorporate existing programmatic tools from open-sourced frameworks, particularly leveraging implementations from AgentS2 \citep{agentS2} and AgentStore \citep{jia2024agentstore}. These tools transform complex GUI sequences into efficient programmatic calls. For example, this AgentS2 tool for spreadsheet manipulation replaces dozens of manual clicks with a single function:

\begin{lstlisting}[language=Python]
def set_cell_values(self, cell_values: dict, app_name: str, sheet_name: str):
    """Set multiple cell values in a spreadsheet.
    Args: cell_values: {"A2": "hello", "B3": 123.45}"""
    return SET_CELL_VALUES_CMD.format(
        cell_values=cell_values, app_name=app_name, sheet_name=sheet_name)
\end{lstlisting}

\textbf{Automatic Scaling with Coding Agents.} Inspired by CoACT-1 \citep{song2025coact}, we adopt the multi-agent paradigm where an orchestrator dynamically delegates subtasks to either a GUI operator or a coding agent that executes Python/Bash scripts. This allows bypassing inefficient GUI sequences through direct programmatic execution. We extend this by mining the coding agent's trajectories for reusable tools: when the coding agent solves subtasks programmatically, we employ an automatic LLM workflow to extract and refine these solutions into parameterized functions, with reflection steps and automated unit testing to ensure correctness. For example, from a trajectory where the coding agent modifies VS Code settings via script, we extract:
\begin{lstlisting}[language=Python]
def add_vs_code_keybinding(key: str, command: str, when: str = ""):
    """Create or update a VS Code keybinding.
    Args: key: "ctrl+j", command: "workbench.action.focusActiveEditorGroup"
    Returns: {"path": "...", "action": "added", "backup": "..."}"""
\end{lstlisting}

\subsection{Synthetic Data Engine for Hybrid Action Tasks}

Large-scale synthetic training tasks for CUAs remain scarce, while existing resources are primarily test sets or complete trajectories with limited reproducibility. To address this, we developed a \textbf{dual-pipeline synthetic data engine} producing 17,000+ verifiable tasks for real-world computer use. Our engine operates through two complementary strategies: \textbf{evaluator-first} generation, ensuring verifiability, and \textbf{instruction-first} generation, creating contextually relevant tasks with diversity.

\subsubsection{Evaluator-First Generation}
This approach begins by collecting state-checking evaluators from computer environments—scripts that verify specific system states (e.g., file existence, application settings, UI elements). We use the atomic evaluator functions in OSWorld~\citep{xie2024osworld} to reprogram these evaluators by modifying parameters and compose multiple evaluators to create complex verification conditions. For example, combining a file-checker with a URL-checker validator creates a task requiring both file manipulation and browsing interaction.

Given these evaluator configurations, we prompt LLMs to generate corresponding tasks that would satisfy the verification conditions. For instance, the file-URL checker combination might generate tasks like ``Navigate to the Python documentation page and download the PDF tutorial to your Documents folder,'' which requires both web browsing to reach the correct URL and file system operations to verify the download. This ensures every generated task has a programmatic way to verify completion, critical for providing clear reward signals during RL training. This approach produces 4,000+ high-quality tasks with guaranteed verifiability.

\subsubsection{Instruction-First Generation}
Following \cite{ramrakhya2025scaling}, this approach generates tasks based on observed system states. Agents explore computer environments through exploratory walks, reaching diverse UI states. At each state, we analyze the current interface and generate contextually appropriate tasks (e.g., ``create a new spreadsheet'' when in a file manager). Task completion is verified by an evaluator agent rather than predefined scripts, allowing flexibility in execution paths. This approach generates 12,000+ tasks that naturally arise from real usage patterns, complementing the systematic coverage of evaluator-first generation.

\subsubsection{Workspace Simulation}
A realistic workspace is crucial for generating meaningful tasks. When synthetic tasks require interaction with specific content, our pipeline triggers a content preparation workflow tailored to task requirements. For example, for code-related tasks, we fetch files from popular GitHub repositories—extracting Python scripts from Hugging Face repos or configuration files from trending projects. For image tasks, we retrieve open-source images from Wikipedia Commons matching relevant categories. For document editing, we generate synthetic documents via LLMs with task-appropriate content.
This targeted approach ensures realistic task contexts: image editing tasks receive actual photos, code refactoring tasks get real implementations, and document tasks operate on properly formatted files. By matching content types to task requirements, we create scenarios that accurately reflect real-world computer use.

\subsubsection{Complementary Design Rationale}
In general, the two approaches serve distinct purposes. Evaluator-first generation produces complex, verifiable tasks ideal for RL training—code-based evaluators provide precise rewards without expensive trajectory verification. However, these tasks tend to be challenging due to evaluators' design and multi-evaluator compositions. Instruction-first generation offers greater diversity through environment exploration, covering more real-world scenarios with naturally easier tasks. This complementary design ensures both reliable RL signals and broad task coverage. We further summarize detailed data statistics in Table~\ref{tab:data_comparison}.

\begin{table}[t]
\centering
\caption{Comparison of our two synthetic data generation strategies.}
\label{tab:data_comparison}
\resizebox{\textwidth}{!}{
\begin{tabular}{lcccccc}
\toprule
\textbf{Strategy} & \textbf{Task Count} & \textbf{Rollout SR (\%)} & \textbf{Avg. Difficulty} & \textbf{Avg. Steps} & \textbf{Total Samples} & \textbf{Total Traj.} \\
\midrule
Evaluator-First & 4K & 29 & Medium-Hard & 6.8 & 33K & 4.8K \\
Instruction-First & 13K & 45 & Easy-Medium & 6.5 & 149K & 22K \\
\bottomrule
\end{tabular}
}
\end{table}

\subsection{Training a Foundation Agent with Hybrid Action}

We train our foundation model using a two-stage approach: supervised fine-tuning on high-quality trajectory demonstrations followed by online reinforcement learning. This curriculum first establishes competency in hybrid action, then optimizes action selection between GUI primitives and programmatic interfaces through self-play.

\subsubsection{Multi-Agent Rollout for Trajectory Generation}

To generate high-quality training data, we deploy a multi-agent system comprising a Planner agent and a specialized Grounder agent. We use OpenAI o3 as the Planner, which operates in a ReAct framework \citep{yao2023react} with Agent-S2-style prompting \citep{agentS2} to enhance reasoning capabilities. The Planner strategically chooses between programmatic calls and GUI actions based on task context and available tools. When GUI interaction is needed, we employ GTA1-7B \citep{yang2025gta1} as the Grounding agent for precise visual localization, ensuring accurate element targeting in complex interfaces.
For each synthetic task, we expose relevant programmatic tools to the Planner and perform 8 rollouts to capture diverse solution strategies. This process generates 26.8K successful trajectories demonstrating effective hybrid action strategies across our synthetic tasks.

\subsubsection{Working Memory Mechanism}
\begin{wrapfigure}{r}{0.53\textwidth}
\vspace{-10pt}
\lstset{basicstyle=\footnotesize\ttfamily}
\begin{lstlisting}
<memory>
Task: Create folder 'Favorites' on bookmarks bar. 
Progress: Chrome open, bookmarks bar visible.
Next: Access bookmark manager via Ctrl+Shift+O.
</memory>
\end{lstlisting}
\vspace{-10pt}
\end{wrapfigure}
Complex hybrid execution paths risk losing context as agents alternate between programmatic tools and GUI actions. We address this through an integrated working memory system using \texttt{<memory></memory>} tags, inspired by~\cite{bonatti2024windows}. The agent autonomously maintains this memory—recording completed steps, extracted values, and intermediate results—ensuring coherent execution without external storage. The common memory content includes: (1) task objectives and constraints, (2) progress tracking across completed actions, and (3) information that must persist across steps (e.g., file paths, UI element states, intermediate values).
For example, during a bookmark management task, the agent maintains structured state information as shown. This mechanism proves crucial for multi-step tasks requiring information persistence across action modality switches.

\subsubsection{Stage 1: Supervised Fine-Tuning}

We fine-tune multiple base models, including UI-TARS-1.5 (7B)~\citep{uitars} and OpenCUA (32B)~\citep{wang2025opencua} on the 26.8K successful trajectories from the rollout system. To ensure balanced training across all trajectory steps, we create individual samples from each turn: for the $i$-th turn, we include messages up to that point but apply loss only to the $i$-th assistant response. This prevents overfitting to early trajectory steps while ensuring each action decision receives equal training weight, teaching the model proper hybrid action at every step of task execution.

\subsubsection{Stage 2: Online Reinforcement Learning}

While SFT provides behavioral foundations, mastering strategic action selection requires learning from exploration. The hybrid action space creates numerous solution paths for each task—some efficient, others suboptimal. Through online RL, agents can discover these optimal strategies via self-play.

We begin by filtering our evaluator-first tasks (4,000+) through 8 rollouts per task with the SFT model, identifying 1,000 tasks where the model succeeds at least once. We define task difficulty as the average success rate across these 8 rollouts. During training, we randomly sample tasks with difficulty scores in $[0.4, 0.8]$—avoiding tasks that are too easy or too challenging to maximize learning efficiency within the model's zone of proximal development.

For policy optimization, we employ a variant of GRPO \citep{shao2024deepseekmath} inspired by DAPO \citep{yu2025dapo}, with key modifications for our hybrid action setting. We remove KL regularization and implement a clip-higher strategy to encourage exploration of diverse action sequences.

To prevent regression toward GUI-only solutions, we design a reward function that incentivizes tool usage. The total reward for a trajectory $\tau$ is:
\begin{equation}
R(\tau) = R_{\text{env}}(\tau) + R_{\text{tool}}(\tau)\,,
\end{equation}
where $R_{\text{env}}(\tau) \in \{-1, 1\}$ is the sparse environment reward (1 for task success, -1 for failure), and the tool-use reward is defined as:
\begin{equation}
R_{\text{tool}}(\tau) = \begin{cases}
0.3 & \text{if } R_{\text{env}}(\tau) = 1 \text{ and } \tau \text{ contains tool calls}, \\
0 & \text{otherwise}.
\end{cases}
\end{equation}
This reward structure teaches the agent not just to succeed, but to succeed efficiently through strategic hybrid action. Notably, we exclude format rewards despite their common use in RL with LLMs. We found in empirical analysis that models struggle with complex tool syntax early in training, causing format penalties to dominate the learning signal and discourage outcome-based learning. By focusing solely on outcome and tool-use rewards, we enable the model to gradually master tool syntax through successful examples rather than punishment, leading to more robust learning. We propagate rewards to each action step and normalize by trajectory length for stable optimization.

%% file: exp.tex
\section{Experiments}
\subsection{Experimental Setup}

\begin{table}[!t]
\caption{Comparison of the state-of-the-art methods on the OSWorld benchmark. We split the results by steps and show the approach type in the second column. We report the success rate (\%) as the evaluation metric in the fourth column. $\dagger$ denotes our reproduced results, averaged across 4 independent runs. Same-colored rows share the same base model.
}
%\vspace{-0.1in}
\resizebox{\textwidth}{!}{
\setlength{\tabcolsep}{6pt}
    \centering
    \begin{tabular}{lcccc}
        \toprule
        \multirow{2}{*}{\textbf{Agent Method}} & \multirow{2}{*}{\textbf{Model Category}} & \multirow{2}{*}{\textbf{Open-Source}} & \multicolumn{2}{c}{\textbf{Success Rate (\%)}} \\
        \cmidrule(lr){4-5}
        & & & \textbf{Max Steps: 15} & \textbf{Max Steps: 50} \\
        \midrule
        o3 \citep{openAI_o3_o4_mini} & General Model & \ding{55} & 9.1 & 17.2 \\
        Claude 3.7 Sonnet \citep{claude37} & General Model & \ding{55} & 27.1 & 35.8 \\
        OpenAI CUA \citep{openAI_o3_o4_mini} & Agentic Model & \ding{55} & 26.0 & 31.3 \\
        Jedi-7B w/ GPT-4o \citep{jedi} & Multi-Agent Framework & \ding{51} & 26.8 & 27.0 \\
        Agent S2 \citep{agentS2} & Multi-Agent Framework & \ding{51} & 27.0 & 34.5 \\
        Qwen2.5-VL-72B \citep{bai2025qwen2} & General Model & \ding{51} & 4.4 & -- \\
        UI-TARS-72B-DPO \citep{uitars} & Agentic Model & \ding{51} & 24.0 & 25.8 \\
        OpenCUA-7B \citep{wang2025opencua} & Agentic Model & \ding{51} & 24.3 & 28.2 \\
        UI-TARS-1.5-7B \citep{uitars} & Agentic Model & \ding{51} & 24.5 & 27.3 \\
        OpenCUA-32B \citep{wang2025opencua} & Agentic Model & \ding{51} & 29.7 & 34.1 \\
        \rowcolor{green!10}
        UI-TARS-1.5-7B$^\dagger$ \citep{uitars} & Agentic Model & \ding{51} & 23.4$^\dagger$ & 26.1$^\dagger$  \\
        \rowcolor{cyan!15}
        OpenCUA-32B$^\dagger$ \citep{wang2025opencua} & Agentic Model & \ding{51} & 33.3$^\dagger$ & 35.1$^\dagger$ \\
        \midrule
        \rowcolor{green!10}
        \textbf{\model-7B-SFT} & Agentic Model & \ding{51} & \textbf{27.0}$^\dagger$ & \textbf{28.5}$^\dagger$ \\
        \rowcolor{green!10}
        \textbf{\model-7B-RL} & Agentic Model & \ding{51} & \textbf{28.9}$^\dagger$ & \textbf{30.2}$^\dagger$ \\
        \rowcolor{cyan!15}
        \textbf{\model-32B-SFT} & Agentic Model & \ding{51} & \textbf{39.0}$^\dagger$ & \textbf{41.5}$^\dagger$ \\
        \rowcolor{cyan!15}
        \textbf{\model-32B-RL} & Agentic Model & \ding{51} & \textbf{41.0}$^\dagger$ & \textbf{43.7}$^\dagger$ \\
        \bottomrule
    \end{tabular}
}
    \label{tab:osworld_comparison}
    \vspace{-0.1in}
\end{table}

\paratitle{Benchmarks.}
We use \textbf{OSWorld-Verified} \citep{xie2024osworld} as our primary benchmark. It is a realistic benchmark featuring a Ubuntu Desktop environment accessible through screen observations, comprising 369 tasks. OSWorld contains diverse tasks spanning common office suites, IDEs, and web browsers, designed to rigorously test an agent's long-horizon planning and visual grounding abilities. Each task is self-contained with a deterministic starting state, a natural language goal, and an automated rule-based evaluator, ensuring reproducible and reliable assessment. 
To evaluate cross-platform generalization, we also test on \textbf{WindowsAgentArena}~\citep{bonatti2024windows}, which contains 154 real-world tasks in Windows 11 environments. This provides an out-of-domain evaluation since our models are primarily trained on Ubuntu-based tasks, testing the transferability of learned hybrid action strategies across operating systems.

\paratitle{Baselines.}
To demonstrate the effectiveness of our approach, we compare our final model against several strong baselines that isolate different components of the agent's capabilities.

\begin{itemize}[leftmargin=*]
    \item \textbf{General Models:} powerful, pre-trained vision-language models that are not specifically fine-tuned for GUI automation. We include leading models like Claude~\citep{claude37} and o3~\citep{openAI_o3_o4_mini} to establish a baseline for generalist, out-of-the-box performance.
    \item \textbf{Multi-Agent Frameworks:} systems that orchestrate multiple components to solve computer-use tasks. These frameworks typically employ a planner-grounder architecture and may be enhanced with additional capabilities such as memory, experience replay, or the integration of a coding agent. Prominent examples include Agent-S2~\citep{agentS2} and Jedi-7B~\citep{jedi}.
    \item \textbf{Specialized Agentic Models:} models that have been specifically fine-tuned or purpose-built for computer control and GUI-centric scenarios. This includes models like OpenAI CUA~\citep{openAI_o3_o4_mini}, UI-TARS~\citep{uitars} and OpenCUA~\citep{wang2025opencua}, which are trained on large datasets of computer interaction trajectories to specialize their abilities for this domain. 
\end{itemize}

\paratitle{Training Details.} Our models are fine-tuned for 3 epochs during the SFT stage with a learning rate of 2e-5. For the SFT stage, we sample 66K steps from trajectories with evaluator-first and instruction-first synthetic data, each 33K. The subsequent online RL stage is trained for 150 steps with a learning rate of 1e-6. All experiments are conducted on NVIDIA H100 GPUs. During training, we control the number of programmatic tools to limit the context length to 32K. 

\paratitle{Evaluation Metrics.} We use the following metrics to measure effectiveness and efficiency:
\textbf{1) Success Rate (SR):} Our primary metric. It is the percentage of tasks the agent successfully completes in a single attempt, as verified by the benchmark's automated evaluators.
\textbf{2) Pass@4:} To account for the stochastic nature of LLM inference, we also report Pass@4. A task is marked as successful under this metric if the agent completes it correctly in at least one of four independent rollout attempts.
\textbf{3) Trajectory Efficiency:} We measure the number of steps an agent takes to complete a task successfully. Each step is either a GUI action or a programmatic tool call. A lower step count indicates higher efficiency.

\begin{table}[t]
\centering
\caption{Detailed performance comparison of agent methods across different domains on OSWorld, measured by success rate (\%) under the 15-step setting. Highlighted cells denote best results for each domain.}
\label{tab:detailed_osw}
\resizebox{\linewidth}{!}{
\begin{tabular}{l c c c c c c c c c c c}
\toprule
\multirow{2}{*}{\textbf{Methods}} & \textbf{Chrome} & \textbf{Calc} & \textbf{Impress} & \textbf{Writer} & \textbf{GIMP} & \textbf{VSCode} & \textbf{Multi Apps} & \textbf{Thunderbird} & \textbf{OS} & \textbf{VLC} & \multirow{2}{*}{\textbf{Avg.}} \\
& \textbf{(46)} & \textbf{(47)} & \textbf{(47)} & \textbf{(23)} & \textbf{(25)} & \textbf{(23)} & \textbf{(101)} & \textbf{(15)} & \textbf{(24)} & \textbf{(17)} & \\
\textit{(avg. runs)} & & & & & & & & & & & \\
\midrule
\rowcolor{green!10}
UI-TARS-1.5-7B & 32.5 & 8.4 & 22.8 & 36.9 & 48.5 & 43.5 & 7.1 & 38.3 & 34.1 & 27.0 & 23.4 \\
\rowcolor{green!10}
& \textit{\small (4.0)} & \textit{\small (4.0)} & \textit{\small (4.0)} & \textit{\small (4.0)} & \textit{\small (4.0)} & \textit{\small (4.0)} & \textit{\small (4.0)} & \textit{\small (4.0)} & \textit{\small (4.0)} & \textit{\small (4.0)} & \textit{\small (4.0)} \\
\rowcolor{green!10}
\model-7B-SFT & 40.7 & 13.3 & 26.9 & 46.7 & \cellcolor{green!30}57.7 & 34.8 & 8.4 & 41.1 & 36.0 & 30.3 & 27.0 \\
\rowcolor{green!10}
& \textit{\small (4.0)} & \textit{\small (4.0)} & \textit{\small (4.0)} & \textit{\small (4.0)} & \textit{\small (4.0)} & \textit{\small (4.0)} & \textit{\small (4.0)} & \textit{\small (4.0)} & \textit{\small (4.0)} & \textit{\small (4.0)} & \textit{\small (4.0)} \\
\rowcolor{green!10}
\model-7B-RL & \cellcolor{green!30}41.2 & \cellcolor{green!30}13.9 & \cellcolor{green!30}27.1 & \cellcolor{green!30}55.4 & 50.0 & \cellcolor{green!30}46.7 & \cellcolor{green!30}10.6 & \cellcolor{green!30}43.3 & \cellcolor{green!30}37.0 & \cellcolor{green!30}33.6 & \cellcolor{green!30}28.9 \\
\rowcolor{green!10}
& \textit{\small (4.0)} & \textit{\small (4.0)} & \textit{\small (4.0)} & \textit{\small (4.0)} & \textit{\small (4.0)} & \textit{\small (4.0)} & \textit{\small (4.0)} & \textit{\small (4.0)} & \textit{\small (4.0)} & \textit{\small (4.0)} & \textit{\small (4.0)} \\
\rowcolor{cyan!15}
OpenCUA-32B & \cellcolor{cyan!35}42.4 & 21.3 & 36.8 & 42.4 & 53.8 & 39.1 & 9.9 & 50.0 & 42.9 & 23.5 & 33.3 \\
\rowcolor{cyan!15}
& \textit{\small (4.0)} & \textit{\small (2.1)} & \textit{\small (2.2)} & \textit{\small (2.2)} & \textit{\small (4.0)} & \textit{\small (3.0)} & \textit{\small (1.9)} & \textit{\small (2.0)} & \textit{\small (1.8)} & \textit{\small (1.8)} & \textit{\small (2.5)} \\
\rowcolor{cyan!15}
\model-32B-SFT & 38.9 & 20.0 & \cellcolor{cyan!35}40.2 & 59.3 & \cellcolor{cyan!35}71.4 & \cellcolor{cyan!35}57.9 & 9.9 & 69.6 & 62.1 & 31.5 & 39.0 \\
\rowcolor{cyan!15}
& \textit{\small (4.0)} & \textit{\small (2.1)} & \textit{\small (2.2)} & \textit{\small (2.2)} & \textit{\small (4.0)} & \textit{\small (3.0)} & \textit{\small (1.9)} & \textit{\small (2.0)} & \textit{\small (1.8)} & \textit{\small (1.8)} & \textit{\small (2.5)} \\
\rowcolor{cyan!15}
\model-32B-RL & 40.6 & \cellcolor{cyan!35}25.7 & \cellcolor{cyan!35}40.2 & \cellcolor{cyan!35}62.5 & 70.0 & 54.3 & \cellcolor{cyan!35}14.9 & \cellcolor{cyan!35}72.1 & \cellcolor{cyan!35}64.0 & \cellcolor{cyan!35}33.3 & \cellcolor{cyan!35}41.0 \\
\rowcolor{cyan!15}
& \textit{\small (4.0)} & \textit{\small (2.1)} & \textit{\small (2.2)} & \textit{\small (2.2)} & \textit{\small (4.0)} & \textit{\small (3.0)} & \textit{\small (1.9)} & \textit{\small (2.0)} & \textit{\small (1.8)} & \textit{\small (1.8)} & \textit{\small (2.5)} \\
\bottomrule
\end{tabular}
}
\end{table}

\subsection{Main Results}

\paratitle{OSWorld Evaluation.}
Table~\ref{tab:osworld_comparison} presents comprehensive results on the OSWorld benchmark across different step budgets. Our \model-7B achieves 28.9\% success rate at 15 steps, surpassing all comparable 7B models, including the strong UI-TARS-1.5-7B baseline (23.4\%) with a 23.5\% relative improvement. More remarkably, \model-32B reaches 41.0\% success rate, outperforming even closed-source systems like Claude 3.7 Sonnet (27.1\%) and OpenAI CUA (26.0\%). We also present the detailed evaluation results per domain in Table~\ref{tab:detailed_osw}. Note that for OpenCUA-series models, due to the sub-optimal speed and infrastructure for inference, the overall average run is less than 4. 

The results validate our hybrid action approach across model scales. While general-purpose models struggle without specialized training (e.g., Qwen2.5-VL-72B at 4.4\% despite 72B parameters), our models achieve superior performance through strategic integration of programmatic tool calls. The consistent improvements from base models (UI-TARS-1.5-7B→\model-7B: +23.5\%, OpenCUA-32B→\model-32B: +23.1\%) demonstrate that hybrid action provides orthogonal benefits to agent capabilities.

\setlength\intextsep{0pt}
\begin{wraptable}{r}{0.4\textwidth}
    \centering
    \caption{Out-of-domain evaluation on WindowsAgentArena.}
    \label{tab:windowsagentarena_simple}
    \resizebox{0.38\textwidth}{!}{
    \begin{tabular}{lc}
    \toprule
    \textbf{Model} & \textbf{SR (\%)} \\
    \midrule
    Qwen2-VL-7B (w/ OpenCUA Data) & 13.5 \\
    UI-TARS-1.5-7B & 18.1 \\
    \textbf{UltraCUA-7B} & \textbf{21.7} \\
    \bottomrule
    \end{tabular}
    }
\end{wraptable}
\paratitle{Cross-Platform Generalization.}
To assess generalization beyond the training domain, we evaluate on WindowsAgentArena without any Windows-specific fine-tuning. Table~\ref{tab:windowsagentarena_simple} shows that \model-7B achieves 21.7\% success rate, outperforming both Qwen2-VL-7B trained with OpenCUA's Windows data (13.5\%) and UI-TARS-1.5-7B (18.1\%). This 20\% relative improvement over UI-TARS demonstrates that hybrid action strategies learned on Ubuntu effectively transfer to Windows environments, validating the domain-agnostic nature of our approach.

\subsection{Ablation Studies}
We conduct a series of ablation studies to dissect our framework and validate the contribution of its key components. These experiments isolate the impact of the hybrid action space, working memory, and reinforcement learning stage on agent performance.

\subsubsection{The Impact of Hybrid Action}
To validate the effectiveness of hybrid action, we examine its impact on both specialized agentic models and powerful multi-agent frameworks.

\paratitle{Impact on Specialized Models.} We compare three configurations: (1) UI-TARS-1.5-7B (GUI-only baseline), (2) our model with tools disabled (\model-7B w/o Tools), and (3) our full model with hybrid action. Table~\ref{tab:hybrid_ablation} shows that hybrid action yields substantial improvements: success rate increases from 23.4\% to 27.0\% (+15.4\% relative) while maintaining similar step counts. The addition of programmatic tools proves essential for effectiveness in complex automation tasks.

\paratitle{Impact on Multi-Agent Frameworks.} To test whether hybrid action benefits extend to state-of-the-art systems, we evaluate our GTA1-7B + o3 rollout framework with and without programmatic tools. As shown in Table~\ref{tab:hybrid_ablation}, hybrid action provides even larger gains in this setting: success rate improves from 44.0\% to 48.2\% (+9.5\% relative) and average steps decrease by 14.9\%. This demonstrates that hybrid action becomes increasingly valuable as the underlying system becomes more capable.

\begin{table}[t!]
\centering
\caption{Impact of hybrid action on different agent architectures. Hybrid action benefits both specialized models and multi-agent frameworks.}
%\vspace{-0.1in}
\label{tab:hybrid_ablation}
\resizebox{0.8\textwidth}{!}{
\begin{tabular}{lccc}
\toprule
\textbf{Model Configuration} & \textbf{Success Rate (\%)} & \textbf{Pass@4} & \textbf{Avg. Steps} \\
\midrule
\rowcolor{gray!15}
\multicolumn{4}{l}{\textit{Agentic Models (Max Steps: 15)}} \\
UI-TARS-1.5-7B (GUI-Only) & 23.4 & 33.3 & 9.31 \\
\model-7B-SFT w/o Tools (GUI-Only) & 25.1 & 34.3 & 9.24 \\
\model-7B-SFT (Hybrid Action) & \textbf{27.0} & \textbf{37.9} & \textbf{8.46} \\
\midrule
\rowcolor{gray!15}
\multicolumn{4}{l}{\textit{Commercial Models \& Multi-Agent Framework (Max Steps: 50)}} \\
Claude-4-Sonnet & 43.9 & -- & -- \\
GTA1-7B + o3 w/o Tools & 44.0 & 60.5 & 15.53 \\
GTA1-7B + o3 (Hybrid Action) & \textbf{48.2} & \textbf{62.4} & \textbf{13.22} \\
\bottomrule
\end{tabular}
}
\vspace{-0.15in}
\end{table}

\subsubsection{The Importance of Reinforcement Learning}
\begin{figure}[t!]
% \vspace{-0.1in}
    \centering
    \begin{subfigure}[b]{0.3\textwidth}
        \centering
        \includegraphics[width=\textwidth]{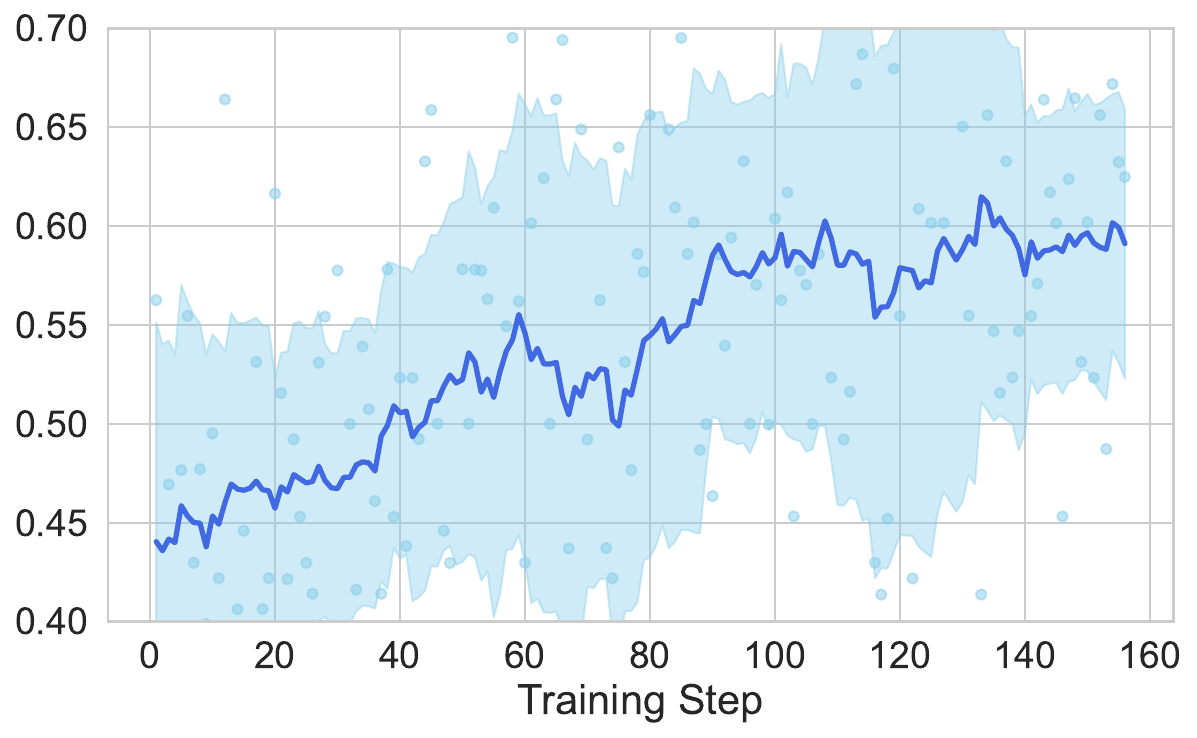}
        \caption{Outcome Reward}
        \label{fig:sub1}
    \end{subfigure}
    % \hfill 
    \begin{subfigure}[b]{0.3\textwidth}
        \centering
        \includegraphics[width=\textwidth]{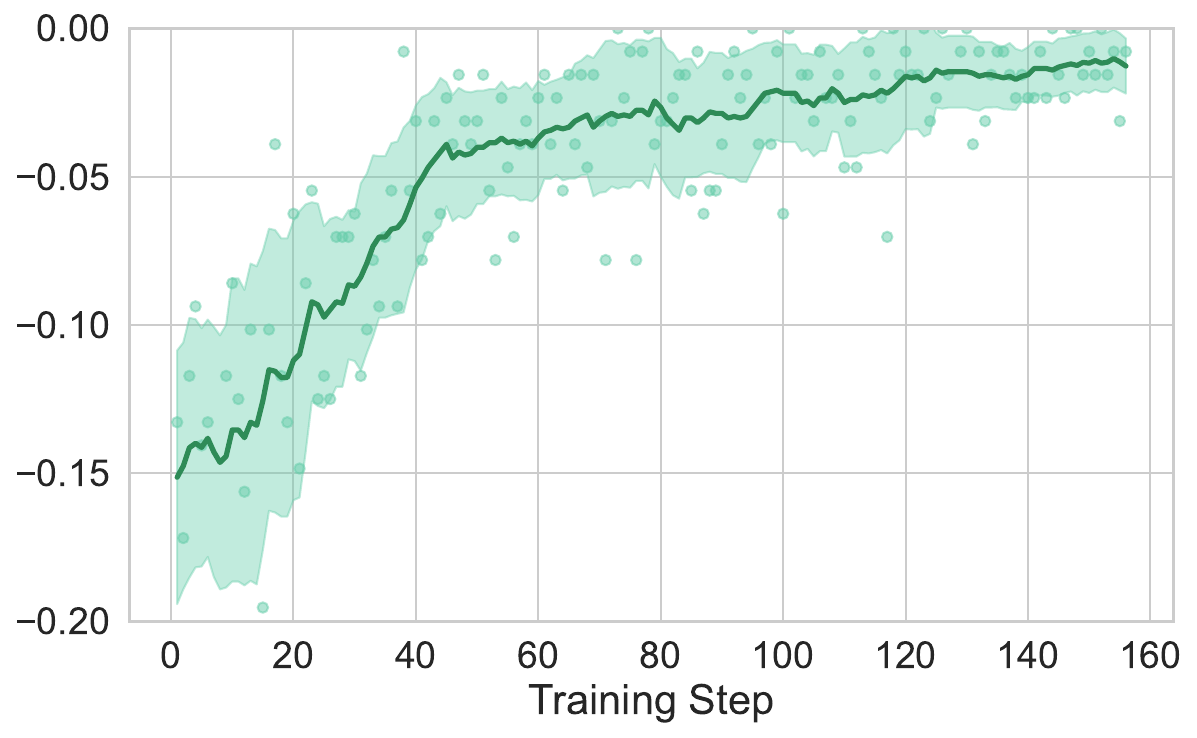}
        \caption{Format Reward}
        \label{fig:sub2}
    \end{subfigure}
    \hfill 
    \begin{subfigure}[b]{0.33\textwidth}
        \centering
        \includegraphics[width=\textwidth]{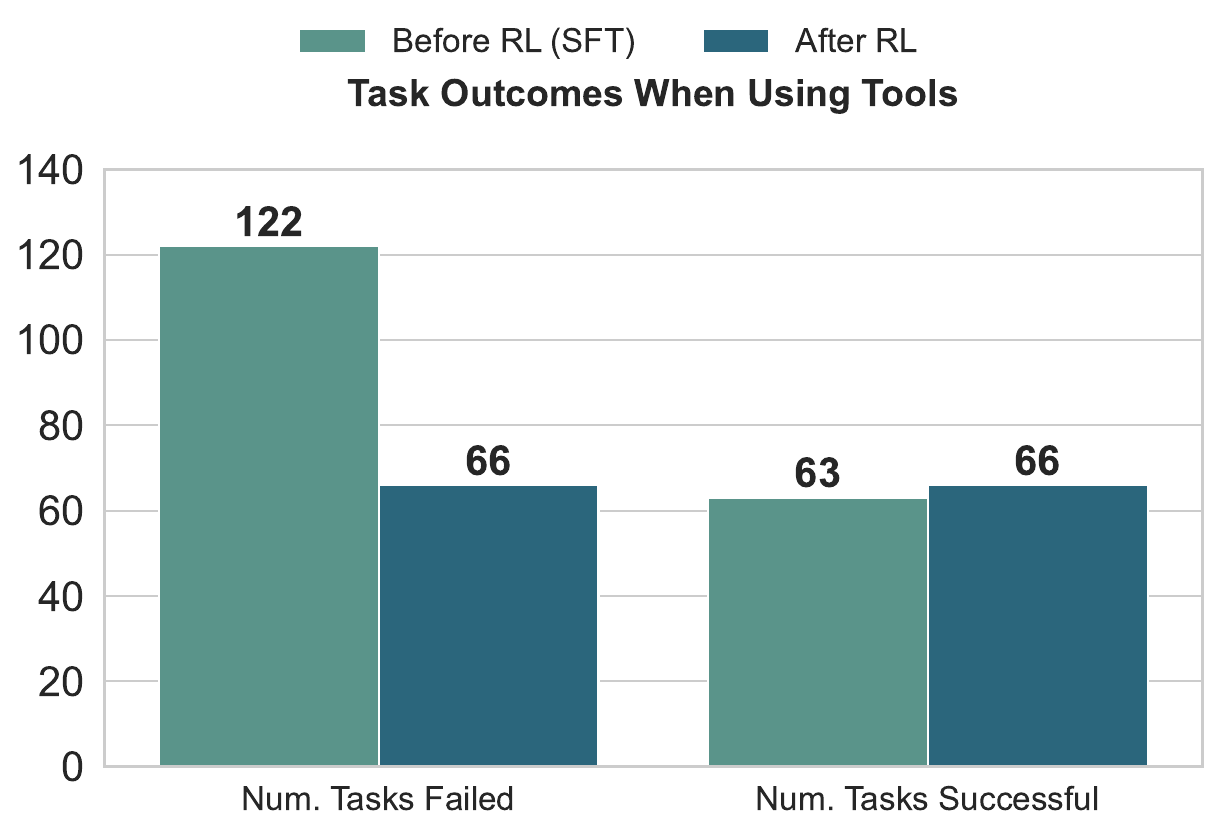}
        \caption{Tool-call Pattern}
        \label{fig:sub3}
    \end{subfigure}
    %\vspace{-0.1in}
    \caption{Evolution of the agent's behavior during reinforcement learning. Rewards are increasing as the number of failed tasks with tool-calls drops after RL.}
    \label{fig:three_figures_total}
    \vspace{-0.15in}
\end{figure}

We evaluate the impact of online RL by comparing models before and after this training stage, for \model-7B. From Table~\ref{tab:osworld_comparison}, we can see that online RL brings 7\% relative improvement (27.0$\rightarrow$28.9). Figure~\ref{fig:three_figures_total} reveals how RL transforms agent behavior in three key ways.
First, outcome rewards increase steadily during RL (Fig.~\ref{fig:sub1}), confirming performance gains. Interestingly, format rewards also improve substantially (Fig.~\ref{fig:sub2}) despite not being explicitly optimized. This suggests agents learn proper tool syntax naturally through successful task completion.
Most significantly, RL reshapes tool-use strategy (Fig.~\ref{fig:sub3}). Tool-related failures drop 46\% (122→66) while successes increase by 5\%, indicating pre-RL models often make harmful tool calls. Correspondingly, overall tool usage decreases, showing agents learn to be selective rather than aggressive with tool deployment.
These results demonstrate that while SFT teaches the mechanics of hybrid action, RL enables strategic decision-making about \textit{when} to use each action type—a crucial distinction for effective automation.

\subsubsection{Impact of Working Memory}

We evaluate working memory by training models with and without \texttt{<memory></memory>} blocks in the SFT data, isolating the contribution of explicit state tracking.
Table~\ref{tab:memory_ablation} shows consistent improvements from working memory: success rate increases from 25.4\% to 27.0\% (+6.3\% relative) and average steps decrease slightly. While modest, these gains are meaningful for tasks requiring persistent state—file operations, form filling, and cross-application workflows. The efficiency improvement suggests memory helps agents avoid redundant actions like re-navigating to previously visited screens or re-extracting obtained information.

\begin{table}[t!]
\centering
\caption{Impact of working memory on model performance. Models are trained with identical data except for the presence of memory blocks.}
%\vspace{-0.1in}
\label{tab:memory_ablation}
\resizebox{0.7\textwidth}{!}{
\begin{tabular}{lccc}
\toprule
\textbf{Model Configuration} & \textbf{Success Rate (\%)} & \textbf{Pass@4} & \textbf{Avg. Steps} \\
\midrule
\model-7B-SFT w/o Memory & 25.4 & 37.1 & 8.56 \\
\model-7B-SFT w/ Memory & \textbf{27.0} & \textbf{37.9} & \textbf{8.46} \\
\midrule
\textit{Relative Improvement} & \textbf{+6.3\%} & \textbf{+2.1\%} & \textbf{-1.2\%} \\
\bottomrule
\end{tabular}
}
\vspace{-0.15in}
\end{table}

\subsection{Analysis on Tool Use Patterns}

To understand how our model leverages the hybrid action space, we analyze tool usage patterns across different application domains and task types.

\begin{wrapfigure}{r}{0.55\textwidth}
% \vspace{-10pt}
\centering
\includegraphics[width=0.54\textwidth]{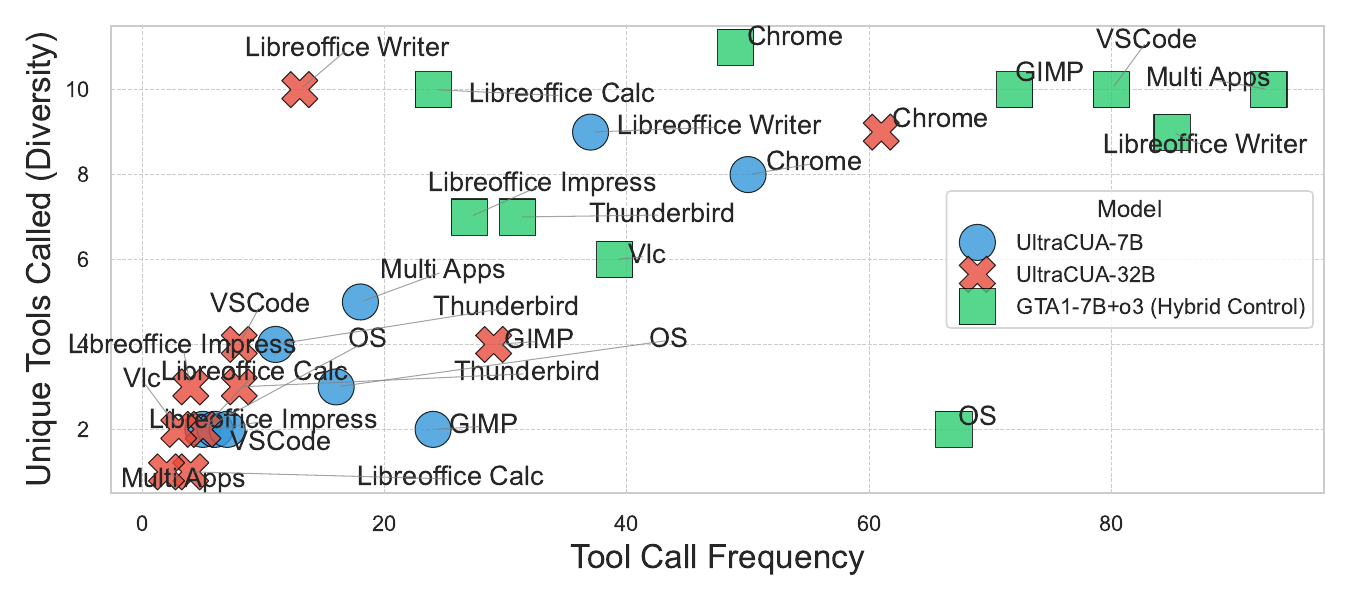}
\vspace{-0.1in}
\caption{Tool-call patterns across domains and models. Stronger models exhibit higher frequency and diversity.}
\label{fig:tool_usage_patterns}
% \vspace{-10pt}
\end{wrapfigure}
\textbf{Tool Usage Scales with Model Capability.} 
Figure~\ref{fig:tool_usage_patterns} reveals a clear correlation between model capability and tool usage sophistication. The multi-agent framework (GTA1-7B+o3) demonstrates extensive tool utilization with 60-80 calls and 8-10 unique tools per domain, while our single models show progressively conservative patterns—UltraCUA-32B uses tools moderately (20-40 calls) and UltraCUA-7B sparingly (0-20 calls).
This pattern validates our hybrid action hypothesis: stronger models not only call tools more frequently but also leverage greater diversity, suggesting they better recognize when programmatic interfaces provide efficiency gains. The trend holds across all domains from office suites to development environments, confirming that effective hybrid action emerges naturally with increased model capability.

\begin{wraptable}{r}{0.45\textwidth}
% \vspace{-10pt}
\centering
\caption{OSWorld with OOD tools.}
% \vspace{-0.1in}
\label{tab:ood_tools}
\resizebox{0.43\textwidth}{!}{
\begin{tabular}{lcc}
\toprule
\textbf{Configuration} & \textbf{SR (\%)} & \textbf{Avg. Steps} \\
\midrule
\model-7B-SFT & 27.0 & 8.46 \\
w/ OOD tools & \textbf{27.5} & 8.80 \\
\bottomrule
\end{tabular}
}
% \vspace{-10pt}
\end{wraptable}
\textbf{Out-of-Distribution Tool Generalization.}
We evaluate the model's ability to utilize tools not seen during training by introducing new programmatic tools at inference time. These tools are unseen during training due to the context length limit.
Table~\ref{tab:ood_tools} shows that models can adapt to unseen tools, achieving modest performance gains (+1.9\% relative SR). However, the increased steps suggest adaptation challenges—models may explore unfamiliar tools before selecting appropriate ones. This zero-shot tool generalization capability also extends beyond single-platform scenarios: Table~\ref{tab:windowsagentarena_simple} demonstrates that our model achieves 21.7\% success rate on Windows tasks despite training exclusively on Ubuntu, outperforming baselines by leveraging its learned hybrid action strategies across platforms and tool ecosystems.

%% file: conclusion.tex
\vspace{-0.1in}
\section{Conclusion}
We introduced \model as foundation CUA models that bridge the critical gap between general-purpose GUI agents and specialized API-based agents. We achieve this through a novel hybrid action space that seamlessly integrates low-level GUI actions with high-level tool use. Our core contributions are a scalable pipeline for automated tool acquisition, a synthetic data engine for generating verifiable hybrid tasks, and a two-stage training curriculum to optimize strategic agent behavior. \model achieves state-of-the-art performance on the real-world benchmarks. Ablation studies confirm that the hybrid action space is the essential driver of this success, demonstrating a novel and more effective paradigm for building strong, robust and efficient agents for general computer control.

%% file: appendix.tex
\appendix
\section{Appendix}
\subsection{Related Work}
\paragraph{Multimodal Agents for Computer Automation.}
The ambition to create agents that can operate GUIs is long-standing, but has seen remarkable progress with the advent of Vision-Language Models (VLMs). Early approaches often relied on structured data like HTML or accessibility trees. More recent and generalizable agents operate directly from pixels and high-level instructions. In web automation, benchmarks like WebArena \citep{zhou2023webarena} and Mind2Web \citep{deng2024mind2web} have driven the development of agents capable of complex online tasks. Similarly, in general computer control, works like CogAgent \citep{hong2024cogagent}, Ferret UI~\cite{you2024ferret, li2024ferret} and OSWorld \citep{xie2024osworld} have demonstrated agents that can navigate desktop environments, and AppAgent \citep{zhang2023appagent} has shown similar capabilities on mobile devices.
Current approaches to GUI automation can be broadly categorized into two paradigms. Multi-Agent systems employ specialized models for different subtasks—for instance, GPT-4o+Aria-UI \citep{yang2024aria} and GTA-1~\citep{yang2025gta1} combine a planner model with a dedicated grounder model, leveraging the strengths of each component for strategic planning and precise visual grounding, respectively. In contrast, Foundation Agent Models like UI-TARS \citep{uitars}, UI-TARS-2~\citep{uitars2}, OpenCUA ~\citep{wang2025opencua} and Ferret-UI Lite~\citep{yang2025ferret} adopt an end-to-end approach, where a single unified model autonomously handles both planning and grounding tasks. While multi-agent systems benefit from modular design and specialized expertise, foundation models offer simpler deployment and potentially better coordination between planning and execution.
A common thread among these powerful agents is their reliance on a primitive action space consisting of clicks, types, and scrolls. While this provides generality, it also leads to the brittleness and long-horizon planning challenges that our work directly addresses. Our contribution is the introduction of a hybrid action space that retains this generality while adding the efficiency and robustness of high-level tools.

\paragraph{Tool and API Augmentation for LLMs.}
Parallel to the development of GUI agents, another line of research has focused on augmenting LLMs with the ability to use external tools and APIs. The seminal work of ToolFormer \citep{schick2023toolformer} showed that models could learn to call APIs to access information they lack. This paradigm was rapidly scaled up by frameworks like ToolLLM \citep{qin2023toolllm} and the Gorilla benchmark \citep{patil2023gorilla}, which enabled models to select from thousands of real-world APIs. Furthermore, the concept of ``tool-making'' \citep{cai2023large} has explored agents that can write their own tools when needed, a capability we incorporate into our tool acquisition pipeline.
Recent advances have introduced reinforcement learning to tool-use training. ReTool \citep{feng2025retool} and ToolRL \citep{qian2025toolrl} pioneered the use of online RL for training end-to-end tool-use agents, demonstrating that reward signals alone can guide models to learn effective tool selection and usage strategies. These methods move beyond supervised learning on static datasets, allowing agents to discover optimal tool-use patterns through interaction and feedback. This RL-based paradigm aligns closely with our approach, where we employ online reinforcement learning to train agents that can strategically alternate between primitive GUI actions and high-level tool calls.
While these tool-augmented agents are highly effective for structured, programmatic tasks, they typically operate in a non-visual, text-based environment and lack the ability to interact with the vast number of applications that do not expose an API. Our work bridges this gap, bringing the power of a rich tool ecosystem to the visually-grounded domain of GUI agents.

\subsection{The Use of Large Language Models}
We used large language models (LLMs) to assist with specific aspects of paper preparation. Specifically, LLMs were employed for: (1) language polishing and grammar checking to improve clarity and readability, (2) formatting suggestions to ensure compliance with conference style guidelines, and (3) recommendations for data visualization approaches to better present experimental results. All research ideas, experimental design, implementation, and core scientific contributions were developed by the authors without LLM assistance. The LLMs served purely as writing and presentation aids.

\subsection{Details for Programmatic Tools}
\label{app:tool}
Table~\ref{tab:tools_overview} summarizes the programmatic tools available across 10 different application domains on OSWorld. The collection comprises 881 tools in total, with individual domains offering between 4 (System) and 135 (VS Code) tools. These tools provide fine-grained control over desktop applications, enabling agents to perform tasks ranging from basic navigation (e.g., \code{jump\_to\_next\_tab}) to complex application-specific operations (e.g., \code{batch\_spreadsheet\_numeric\_formatter}). The comprehensive tool coverage ensures that agents can effectively automate diverse desktop workflows across different software environments.

\vspace{0.2in}
\begin{table}[ht]
\centering
\caption{Overview of available tools across different domains.}
\label{tab:tools_overview}
\small % Make the entire table use smaller font
\renewcommand{\arraystretch}{1.1} % Reduce row spacing for compactness
\begin{tabular}{@{}lc>{\raggedright\arraybackslash}p{0.5\linewidth}@{}}
\toprule
\rowcolor{tableheader}
\textcolor{white}{\textbf{Domain}} & \textcolor{white}{\textbf{Tool Count}} & \textcolor{white}{\textbf{Example Tools}} \\
% \textbf{Domain} & \textbf{\# Tools} & \textbf{Example Tools} \\
\midrule
\rowcolor{gray!5}
Chrome & 69 & \texttt{jump\_to\_next\_tab}, \texttt{chrome\_domain\_data\_wiper}, \texttt{open\_downloads\_page} \\
GIMP & 88 & \texttt{save\_image\_as}, \texttt{undo\_last\_action}, \texttt{swap\_foreground\_background\_colors} \\
\rowcolor{gray!5}
LibreOffice General & 41 & \texttt{open\_find\_and\_replace}, \texttt{open\_print\_preview}, \texttt{open\_hyperlink\_dialog} \\
LibreOffice Calc & 114 & \texttt{spreadsheet\_column\_formula\_injector}, \texttt{batch\_spreadsheet\_numeric\_formatter} \\
\rowcolor{gray!5}
LibreOffice Impress & 75 & \texttt{set\_line\_spacing\_1}, \texttt{insert\_non\_breaking\_space}, \texttt{apply\_subscript} \\
LibreOffice Writer & 123 & \texttt{select\_to\_start\_of\_next\_page}, \texttt{select\_to\_start\_of\_paragraph} \\
\rowcolor{gray!5}
System & 4 & \texttt{open\_system\_terminal\_and\_execute}, \texttt{open\_app\_or\_filename} \\
Thunderbird & 119 & \texttt{open\_message\_in\_conversation}, \texttt{delete\_message\_permanently} \\
\rowcolor{gray!5}
VLC & 83 & \texttt{set\_video\_as\_wallpaper}, \texttt{volume\_up}, \texttt{jump\_1\_minute\_forward} \\
VS Code & 135 & \texttt{add\_vs\_code\_keybinding}, \texttt{vscode\_exclude\_folders}, \texttt{search\_within\_current\_file} \\
\midrule
\textbf{Total} & \textbf{881} & \\
\bottomrule
\end{tabular}
\end{table}

\subsection{Details for Synthetic Tasks}
We generated a comprehensive synthetic dataset of 17,864 tasks across 10 application domains using two complementary approaches. As shown in Table~\ref{tab:synthetic_data_overview}, the evaluator-first approach contributed 4,387 high-quality tasks with complex multi-step instructions, while the instruction-first approach generated 13,477 tasks to ensure broad coverage of application functionalities.

The dataset spans diverse applications from productivity tools (LibreOffice suite with 5,885 combined tasks) to specialized software like GIMP (1,121 tasks) and development environments like VS Code (1,990 tasks). Chrome represents the largest single-domain category with 2,826 tasks, reflecting the importance of web interactions. The multi-apps category (2,113 tasks) specifically tests cross-application workflows. Task complexity varies from simple operations (e.g., \textit{``Change the text alignment to Center''}) to sophisticated procedures requiring multiple coordinated actions (e.g., \textit{``Convert video to MP4 and save with a new filename''}), ensuring comprehensive evaluation of agents' GUI navigation and task execution capabilities.

\begin{table}[htbp]
\centering
\caption{Overview of synthetic data generation across different domains.}
\small
\label{tab:synthetic_data_overview}
\begin{tabular}{>{\raggedright\arraybackslash}p{0.12\linewidth}ccp{0.44\linewidth}c}
\toprule
\rowcolor{tableheader}
\textcolor{white}{\textbf{Domain}} & \textcolor{white}{\textbf{Eval-First}} & \textcolor{white}{\textbf{Instr-First}} & \textcolor{white}{\textbf{Example Instructions}} & \textcolor{white}{\textbf{Total}} \\
\midrule
\rowcolor{gray!5}
Chrome & 751 & 2,075 & \textit{``Find hotels in Paris for 2 adults for three nights starting next Friday and sort the list by lowest price.''} \par\vspace{1pt} \textit{``Restore the previous session pages in Google Chrome.''} & 2,826 \\
GIMP & 401 & 720 & \textit{``Please replace the current white backdrop with a solid green color, but keep the black circle in the centre exactly as it is.''} \par\vspace{1pt} \textit{``In GIMP, navigate to the Display section and set the check style to Medium checks.''} & 1,121 \\
\rowcolor{gray!5}
LibreOffice Calc & 651 & 1,496 & \textit{``Open the spreadsheet and make the entire header row (row 1) bold.''} \par\vspace{1pt} \textit{``Protect the sheet Sheet2 in LibreOffice Calc.''} & 2,147 \\
LibreOffice Impress & 501 & 1,397 & \textit{``Make every slide in this deck use a solid dark-green background (RGB 0 128 0). I'd like all the pages to share that exact colour so the presentation looks consistent.''} \par\vspace{1pt} \textit{``Add a video from /videos/video3.mov to slide 3 in LibreOffice Impress.''} & 1,898 \\
\rowcolor{gray!5}
LibreOffice Writer & 851 & 989 & \textit{``Change the default font in LibreOffice Writer to Calibri.''} \par\vspace{1pt} \textit{``Change the text alignment to Center in LibreOffice Writer.''} & 1,840 \\
OS/System & 301 & 1,197 & \textit{``I accidentally created a file called 'draft.txt' on my Desktop. Please delete it completely so it's no longer there.''} \par\vspace{1pt} \textit{``View the partitioning table of the disk named \{disk\_name\} in the Disks app.''} & 1,498 \\
\rowcolor{gray!5}
Thunderbird & 351 & 1,084 & \textit{``Create a new folder named 'ToSort' inside the Local Folders section.''} \par\vspace{1pt} \textit{``Import contacts from Windows Mail into Thunderbird.''} & 1,435 \\
VLC & 330 & 666 & \textit{``Open the cat photo in VLC and set it as my desktop wallpaper.''} \par\vspace{1pt} \textit{``Play the current video in VLC Media Player.''} & 996 \\
\rowcolor{gray!5}
VS Code & 250 & 1,740 & \textit{``Could you open VS Code and create a new text file named 'meeting\_notes.txt' inside the folder '/home/user/notes'? Make sure to save the file before you finish.''} \par\vspace{1pt} \textit{``Search for the term Data Structure in the document and highlight it in LibreOffice Writer''} & 1,990 \\
Multi-apps & -- & 2,113 & \textit{``Change the desktop wallpaper to Desert on the Ubuntu desktop.''} \par\vspace{1pt} \textit{``Search for JavaScript in Brave settings and enable it.''} & 2,113 \\
\midrule
\textbf{Total} & \textbf{4,387} & \textbf{13,477} & & \textbf{17,864} \\
\bottomrule
\end{tabular}
\end{table}

\subsection{Qualitative Examples}
\begin{figure}[h!]
    \centering
    \includegraphics[width=\linewidth]{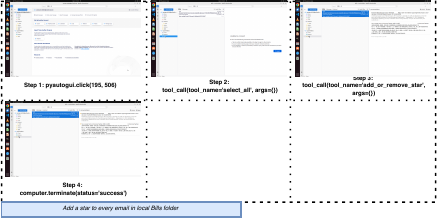}
    % \vspace{-0.3in}
    \caption{An example of \model-32B helping processing emails. The agent alternates between low-level actions and programmatic tool calls smartly, leading to efficient completion of the task.}
    \label{fig:qual_3}
    % \vspace{-0.2in}
\end{figure}

\begin{figure}[h!]
    \centering
    \includegraphics[width=\linewidth]{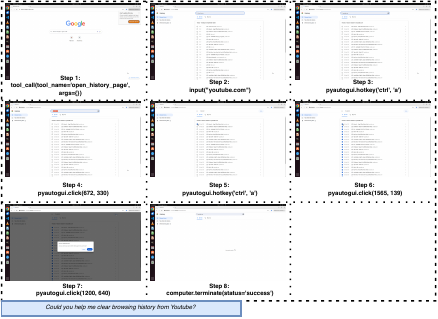}
    % \vspace{-0.3in}
    \caption{An example of \model-32B helping clear certain Chrome history with hybrid action. The agent calls prothe grammatic tool at the first step to assist in going into the desired page directly.}
    \label{fig:qual_1}
    % \vspace{-0.2in}
\end{figure}

\begin{figure}[h!]
    \centering
    \includegraphics[width=\linewidth]{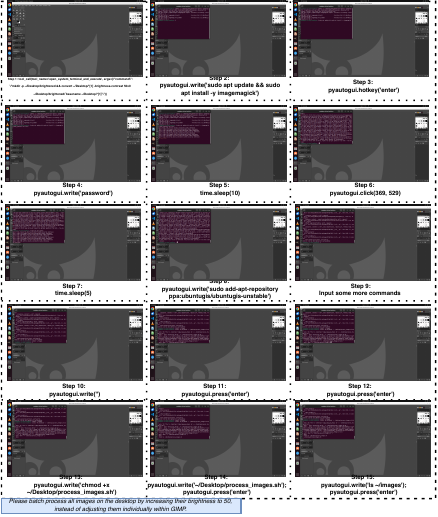}
    % \vspace{-0.3in}
    \caption{An example of \model-32B helping processing images with hybrid action. The model starts coding at the very first step by calling the terminal tool, and finally writes a bash script and executes it to make the task successful.}
    \label{fig:qual_2}
    % \vspace{-0.2in}
\end{figure}

To illustrate the practical advantages of our hybrid action paradigm, we present three representative examples in Figures \ref{fig:qual_3}, \ref{fig:qual_1}, and \ref{fig:qual_2}. These cases highlight how \model strategically selects between high-level programmatic tools and low-level GUI actions to enhance efficiency, tackle complex problems, and ensure robust execution.

First, the email-starring task (Figure \ref{fig:qual_3}) exemplifies the agent's capacity for intelligent and fluid alternation between control modes. The process begins with a precise low-level GUI click to select the target ``Bills'' folder, effectively setting the context. Immediately following this, the agent switches to high-level, reliable tool calls—\texttt{select\_all} and \texttt{add\_or\_remove\_star}—to execute the core bulk operation. This strategic handoff from a specific GUI action to general-purpose tools ensures both precision and operational robustness.

In the second example (Figure \ref{fig:qual_1}), the agent is asked to clear YouTube browsing history. Instead of relying on a potentially brittle sequence of clicks through menus, it initiates the workflow with a single programmatic tool call, \texttt{open\_history\_page}, to navigate directly to the correct settings page. Subsequently, it seamlessly transitions to primitive GUI actions—typing into a search field and clicking buttons—to perform the more nuanced task of filtering and deleting the specific entries. This demonstrates a practical fusion of programmatic speed for navigation and GUI flexibility for manipulation.

Finally, a more complex scenario in Figure \ref{fig:qual_2} showcases the model's ability to automate workflows that are intractable for purely GUI-based agents. When tasked with batch-processing images on the desktop, \model correctly identifies the need for a scripted solution. It programmatically opens a system terminal, installs the necessary software (\texttt{imagemagick}), and proceeds to write and execute a multi-line shell script to automate the entire process. This ability to generate and utilize code represents a significant leap in problem-solving capability.

\vspace{0.2in}

%% file: ref.bib
@article{ramrakhya2025scaling,
  title={Scaling synthetic task generation for agents via exploration},
  author={Ramrakhya, Ram and Szot, Andrew and Attia, Omar and Yang, Yuhao and Nguyen, Anh and Mazoure, Bogdan and Gan, Zhe and Agrawal, Harsh and Toshev, Alexander},
  journal={arXiv preprint arXiv:2509.25047},
  year={2025}
}

@article{wang2025opencua,
  title={Opencua: Open foundations for computer-use agents},
  author={Wang, Xinyuan and Wang, Bowen and Lu, Dunjie and Yang, Junlin and Xie, Tianbao and Wang, Junli and Deng, Jiaqi and Guo, Xiaole and Xu, Yiheng and Wu, Chen Henry and others},
  journal={arXiv preprint arXiv:2508.09123},
  year={2025}
}

@article{yang2024aria,
  title={Aria-ui: Visual grounding for gui instructions},
  author={Yang, Yuhao and Wang, Yue and Li, Dongxu and Luo, Ziyang and Chen, Bei and Huang, Chao and Li, Junnan},
  journal={arXiv preprint arXiv:2412.16256},
  year={2024}
}

@article{feng2025retool,
  title={Retool: Reinforcement learning for strategic tool use in llms},
  author={Feng, Jiazhan and Huang, Shijue and Qu, Xingwei and Zhang, Ge and Qin, Yujia and Zhong, Baoquan and Jiang, Chengquan and Chi, Jinxin and Zhong, Wanjun},
  journal={arXiv preprint arXiv:2504.11536},
  year={2025}
}

@article{qian2025toolrl,
  title={Toolrl: Reward is all tool learning needs},
  author={Qian, Cheng and Acikgoz, Emre Can and He, Qi and Wang, Hongru and Chen, Xiusi and Hakkani-T{\"u}r, Dilek and Tur, Gokhan and Ji, Heng},
  journal={arXiv preprint arXiv:2504.13958},
  year={2025}
}

@article{jia2024agentstore,
  title={Agentstore: Scalable integration of heterogeneous agents as specialized generalist computer assistant},
  author={Jia, Chengyou and Luo, Minnan and Dang, Zhuohang and Sun, Qiushi and Xu, Fangzhi and Hu, Junlin and Xie, Tianbao and Wu, Zhiyong},
  journal={arXiv preprint arXiv:2410.18603},
  year={2024}
}

@article{uitars2,
  title={Ui-tars-2 technical report: Advancing gui agent with multi-turn reinforcement learning},
  author={Wang, Haoming and Zou, Haoyang and Song, Huatong and Feng, Jiazhan and Fang, Junjie and Lu, Junting and Liu, Longxiang and Luo, Qinyu and Liang, Shihao and Huang, Shijue and others},
  journal={arXiv preprint arXiv:2509.02544},
  year={2025}
}

@article{bai2025qwen2,
  title={Qwen2. 5-vl technical report},
  author={Bai, Shuai and Chen, Keqin and Liu, Xuejing and Wang, Jialin and Ge, Wenbin and Song, Sibo and Dang, Kai and Wang, Peng and Wang, Shijie and Tang, Jun and others},
  journal={arXiv preprint arXiv:2502.13923},
  year={2025}
}

@article{bonatti2024windows,
  title={Windows agent arena: Evaluating multi-modal os agents at scale},
  author={Bonatti, Rogerio and Zhao, Dan and Bonacci, Francesco and Dupont, Dillon and Abdali, Sara and Li, Yinheng and Lu, Yadong and Wagle, Justin and Koishida, Kazuhito and Bucker, Arthur and others},
  journal={arXiv preprint arXiv:2409.08264},
  year={2024}
}

@techreport{claude37,
  title={Claude 3.7 Sonnet and Claude Code},
  author={Anthropic},
  year={2025},
  institution={Anthropic},
  url={https://www.anthropic.com/news/claude-3-7-sonnet},
  note={System Card},
  type={Technical Report}
}

@article{jedi,
  title={Scaling Computer-Use Grounding via User Interface Decomposition and Synthesis},
  author={Xie, Tianbao and Deng, Jiaqi and Li, Xiaochuan and Yang, Junlin and Wu, Haoyuan and Chen, Jixuan and Hu, Wenjing and Wang, Xinyuan and Xu, Yuhui and Wang, Zekun and others},
  journal={arXiv preprint arXiv:2505.13227},
  year={2025}
}

@article{yang2025gta1,
  title={Gta1: Gui test-time scaling agent},
  author={Yang, Yan and Li, Dongxu and Dai, Yutong and Yang, Yuhao and Luo, Ziyang and Zhao, Zirui and Hu, Zhiyuan and Huang, Junzhe and Saha, Amrita and Chen, Zeyuan and others},
  journal={arXiv preprint arXiv:2507.05791},
  year={2025}
}

@article{uitars,
  title={UI-TARS: Pioneering Automated GUI Interaction with Native Agents},
  author={Qin, Yujia and Ye, Yining and Fang, Junjie and Wang, Haoming and Liang, Shihao and Tian, Shizuo and Zhang, Junda and Li, Jiahao and Li, Yunxin and Huang, Shijue and others},
  journal={arXiv preprint arXiv:2501.12326},
  year={2025}
}

@techreport{openAI_o3_o4_mini,
  title={OpenAI o3 and o4-mini System Card},
  author={OpenAI},
  year={2025},
  institution={OpenAI},
  url={https://cdn.openai.com/pdf/2221c875-02dc-4789-800b-e7758f3722c1/o3-and-o4-mini-system-card.pdf},
  note={System Card},
  type={Technical Report}
}

@inproceedings{zhou2023webarena,
  title={Webarena: A realistic web environment for building autonomous agents},
  author={Zhou, Shuyan and Xu, Frank F and Zhu, Hao and Zhou, Xuhui and Lo, Robert and Sridhar, Abishek and Cheng, Xianyi and Ou, Tianyue and Bisk, Yonatan and Fried, Daniel and others},
  booktitle={International Conference on Learning Representations},
  volume={2024},
  pages={15585--15606},
  year={2024}
}

@article{deng2024mind2web,
  title={Mind2web: Towards a generalist agent for the web},
  author={Deng, Xiang and Gu, Yu and Zheng, Boyuan and Chen, Shijie and Stevens, Sam and Wang, Boshi and Sun, Huan and Su, Yu},
  journal={Advances in Neural Information Processing Systems},
  volume={36},
  pages={28091--28114},
  year={2023}
}

@inproceedings{hong2024cogagent,
  title={Cogagent: A visual language model for gui agents},
  author={Hong, Wenyi and Wang, Weihan and Lv, Qingsong and Xu, Jiazheng and Yu, Wenmeng and Ji, Junhui and Wang, Yan and Wang, Zihan and Dong, Yuxiao and Ding, Ming and others},
  booktitle={Proceedings of the IEEE/CVF conference on computer vision and pattern recognition},
  pages={14281--14290},
  year={2024}
}

@article{xie2024osworld,
  title={Osworld: Benchmarking multimodal agents for open-ended tasks in real computer environments},
  author={Xie, Tianbao and Zhang, Danyang and Chen, Jixuan and Li, Xiaochuan and Zhao, Siheng and Cao, Ruisheng and Hua, Toh J and Cheng, Zhoujun and Shin, Dongchan and Lei, Fangyu and others},
  journal={Advances in Neural Information Processing Systems},
  volume={37},
  pages={52040--52094},
  year={2024}
}

@inproceedings{zhang2023appagent,
  title={Appagent: Multimodal agents as smartphone users},
  author={Zhang, Chi and Yang, Zhao and Liu, Jiaxuan and Li, Yanda and Han, Yucheng and Chen, Xin and Huang, Zebiao and Fu, Bin and Yu, Gang},
  booktitle={Proceedings of the 2025 CHI Conference on Human Factors in Computing Systems},
  pages={1--20},
  year={2025}
}

@article{schick2023toolformer,
  title={Toolformer: Language models can teach themselves to use tools},
  author={Schick, Timo and Dwivedi-Yu, Jane and Dess{\`\i}, Roberto and Raileanu, Roberta and Lomeli, Maria and Hambro, Eric and Zettlemoyer, Luke and Cancedda, Nicola and Scialom, Thomas},
  journal={Advances in neural information processing systems},
  volume={36},
  pages={68539--68551},
  year={2023}
}

@inproceedings{qin2023toolllm,
  title={Toolllm: Facilitating large language models to master 16000+ real-world apis},
  author={Qin, Yujia and Liang, Shihao and Ye, Yining and Zhu, Kunlun and Yan, Lan and Lu, Yaxi and Lin, Yankai and Cong, Xin and Tang, Xiangru and Qian, Bill and others},
  booktitle={International Conference on Learning Representations},
  volume={2024},
  pages={9695--9717},
  year={2024}
}

@article{patil2023gorilla,
  title={Gorilla: Large Language Model Connected with Massive APIs},
  author={Patil, Shishir G and Zhang, Tianjun and Wang, Xin and Gonzalez, Joseph E},
  journal={arXiv preprint arXiv:2305.15334},
  year={2023}
}

@inproceedings{cai2023large,
  title={Large language models as tool makers},
  author={Cai, Tianle and Wang, Xuezhi and Ma, Tengyu and Chen, Xinyun and Zhou, Denny},
  booktitle={International Conference on Learning Representations},
  volume={2024},
  pages={54067--54089},
  year={2024}
}

@article{song2025coact,
  title={Coact-1: Computer-using agents with coding as actions},
  author={Song, Linxin and Dai, Yutong and Prabhu, Viraj and Zhang, Jieyu and Shi, Taiwei and Li, Li and Li, Junnan and Savarese, Silvio and Chen, Zeyuan and Zhao, Jieyu and others},
  journal={arXiv preprint arXiv:2508.03923},
  year={2025}
}

@article{yao2023react,
  title={React: Synergizing reasoning and acting in language models},
  author={Yao, Shunyu and Zhao, Jeffrey and Yu, Dian and Du, Nan and Shafran, Izhak and Narasimhan, Karthik and Cao, Yuan},
  journal={arXiv preprint arXiv:2210.03629},
  year={2022}
}

@article{shao2024deepseekmath,
  title={Deepseekmath: Pushing the limits of mathematical reasoning in open language models},
  author={Shao, Zhihong and Wang, Peiyi and Zhu, Qihao and Xu, Runxin and Song, Junxiao and Bi, Xiao and Zhang, Haowei and Zhang, Mingchuan and Li, YK and Wu, Yang and others},
  journal={arXiv preprint arXiv:2402.03300},
  year={2024}
}

@article{yu2025dapo,
  title={Dapo: An open-source llm reinforcement learning system at scale},
  author={Yu, Qiying and Zhang, Zheng and Zhu, Ruofei and Yuan, Yufeng and Zuo, Xiaochen and Yue, Yu and Dai, Weinan and Fan, Tiantian and Liu, Gaohong and Liu, Lingjun and others},
  journal={arXiv preprint arXiv:2503.14476},
  year={2025}
}

@article{shaw2024pixels,
  title={From pixels to ui actions: Learning to follow instructions via graphical user interfaces},
  author={Shaw, Peter and Joshi, Mandar and Cohan, James and Berant, Jonathan and Pasupat, Panupong and Hu, Hexiang and Khandelwal, Urvashi and Lee, Kenton and Toutanova, Kristina N},
  journal={Advances in Neural Information Processing Systems},
  volume={36},
  pages={34354--34370},
  year={2023}
}

@inproceedings{rawles2024androidworld,
  title={Androidworld: A dynamic benchmarking environment for autonomous agents},
  author={Rawles, Chris and Clinckemaillie, Sarah and Chang, Yifan and Waltz, Jonathan and Lau, Gabrielle and Fair, Marybeth and Li, Alice and Bishop, William and Li, Wei and Campbell-Ajala, Folawiyo and others},
  booktitle={International Conference on Learning Representations},
  volume={2025},
  pages={406--441},
  year={2025}
}

@inproceedings{koh2024visualwebarena,
  title={Visualwebarena: Evaluating multimodal agents on realistic visual web tasks},
  author={Koh, Jing Yu and Lo, Robert and Jang, Lawrence and Duvvur, Vikram and Lim, Ming and Huang, Po-Yu and Neubig, Graham and Zhou, Shuyan and Salakhutdinov, Russ and Fried, Daniel},
  booktitle={Proceedings of the 62nd Annual Meeting of the Association for Computational Linguistics (Volume 1: Long Papers)},
  pages={881--905},
  year={2024}
}

@article{zheng2024gpt4vision,
  title={Gpt-4v (ision) is a generalist web agent, if grounded},
  author={Zheng, Boyuan and Gou, Boyu and Kil, Jihyung and Sun, Huan and Su, Yu},
  journal={arXiv preprint arXiv:2401.01614},
  year={2024}
}

@article{yan2023gpt4v,
  title={GPT-4V in wonderland: Large multimodal models for zero-shot smartphone GUI navigation},
  author={Yan, An and Yang, Zhengyuan and Zhu, Wanrong and Lin, Kevin and Li, Linjie and Wang, Jianfeng and Yang, Jianwei and Zhong, Yiwu and McAuley, Julian and others},
  journal={arXiv preprint arXiv:2311.07562},
  year={2023}
}

@inproceedings{mialon2024gaia,
  title={GAIA: A benchmark for general AI assistants},
  author={Mialon, Gr{\'e}goire and Fourrier, Cl{\'e}mentine and Swift, Craig and Wolf, Thomas and LeCun, Yann and Scialom, Thomas},
  booktitle={International Conference on Learning Representations (ICLR)},
  year={2024}
}

@article{zhang2025agentorchestra,
  title={Agentorchestra: A hierarchical multi-agent framework for general-purpose task solving},
  author={Zhang, Wentao and Cui, Ce and Zhao, Yilei and Hu, Rui and Liu, Yang and Zhou, Yahui and An, Bo},
  journal={arXiv e-prints},
  pages={arXiv--2506},
  year={2025}
}

@article{qin2023tool,
  title={Tool learning with foundation models},
  author={Qin, Yujia and Hu, Shengding and Lin, Yankai and Chen, Weize and Ding, Ning and Cui, Ganqu and Zeng, Zheni and Zhou, Xuanhe and Huang, Yufei and Xiao, Chaojun and others},
  journal={ACM Computing Surveys},
  volume={57},
  number={4},
  pages={1--40},
  year={2024},
  publisher={ACM New York, NY}
}

@article{tang2023toolalpaca,
  title={Toolalpaca: Generalized tool learning for language models with 3000 simulated cases},
  author={Tang, Qiaoyu and Deng, Ziliang and Lin, Hongyu and Han, Xianpei and Liang, Qiao and Cao, Boxi and Sun, Le},
  journal={arXiv preprint arXiv:2306.05301},
  year={2023}
}

@article{agentS2,
  title={Agent S2: A compositional generalist-specialist framework for computer use agents},
  author={Agashe, Saaket and Wong, Kyle and Tu, Vincent and Yang, Jiachen and Li, Ang and Wang, Xin Eric},
  journal={arXiv preprint arXiv:2504.00906},
  year={2025}
}

@article{li2024ferret,
  title={Ferret-ui 2: Mastering universal user interface understanding across platforms},
  author={Li, Zhangheng and You, Keen and Zhang, Haotian and Feng, Di and Agrawal, Harsh and Li, Xiujun and Moorthy, Mohana Prasad Sathya and Nichols, Jeff and Yang, Yinfei and Gan, Zhe},
  journal={arXiv preprint arXiv:2410.18967},
  year={2024}
}

@inproceedings{you2024ferret,
  title={Ferret-ui: Grounded mobile ui understanding with multimodal llms},
  author={You, Keen and Zhang, Haotian and Schoop, Eldon and Weers, Floris and Swearngin, Amanda and Nichols, Jeffrey and Yang, Yinfei and Gan, Zhe},
  booktitle={European Conference on Computer Vision},
  pages={240--255},
  year={2024},
  organization={Springer}
}

@article{yang2025ferret,
  title={Ferret-UI Lite: Lessons from Building Small On-Device GUI Agents},
  author={Yang, Zhen and Dou, Zi-Yi and Feng, Di and Huang, Forrest and Nguyen, Anh and You, Keen and Attia, Omar and Yang, Yuhao and Feng, Michael and Zhang, Haotian and others},
  journal={arXiv preprint arXiv:2509.26539},
  year={2025}
}
